\pdfoutput=1

\documentclass[11pt]{article}

\usepackage[]{ACL2023}

\usepackage{times}
\usepackage{latexsym}

\usepackage[T1]{fontenc}

\usepackage[utf8]{inputenc}

\usepackage{microtype}

\usepackage{inconsolata}
\usepackage{algorithm}
\usepackage{algorithmic}
\usepackage{subfigure}
\usepackage{makecell}

\usepackage{multirow}
\usepackage{makecell}
\usepackage{subfigure}
\usepackage{multicol}
\usepackage{multirow}
\usepackage{url}
\usepackage{amsmath,amsfonts,amsthm} 
\usepackage{graphicx}
\usepackage{subfigure}
 \usepackage{balance}
\usepackage{color}
\usepackage{tabularx}
\usepackage{array}
\usepackage{booktabs}
\usepackage{amssymb}
\usepackage{bbding}

\title{Multiview Identifiers Enhanced Generative Retrieval}

\author{Yongqi Li$^{1}$, Nan Yang$^{2}$, Liang Wang$^{2}$, Furu Wei$^{2}$, Wenjie Li$^{1}$ \\
        $^{1}$The Hong Kong Polytechnic University 
        $^{2}$Microsoft\\
        \texttt{liyongqi0@gmail.com} \\ 
        \texttt{\{nanya,wangliang,fuwei\}@microsoft.com cswjli@comp.polyu.edu.hk}}

\begin{document}
\maketitle
\begin{abstract}
Instead of simply \textit{matching} a query to pre-existing passages, generative retrieval \textit{generates} identifier strings of passages as the retrieval target. At a cost, the identifier must be distinctive enough to represent a passage. Current approaches use either a numeric ID or a text piece (such as a title or substrings) as the identifier. However, these identifiers cannot cover a passage's content well. As such, we are motivated to propose a new type of identifier, synthetic identifiers, that are generated based on the content of a passage and could integrate contextualized information that text pieces lack. Furthermore, we simultaneously consider multiview identifiers, including synthetic identifiers, titles, and substrings. These views of identifiers complement each other and facilitate the holistic ranking of passages from multiple perspectives. We conduct a series of experiments on three public datasets, and the results indicate that our proposed approach performs the best in generative retrieval, demonstrating its effectiveness and robustness. The code is released at \url{https://github.com/liyongqi67/MINDER}.
\end{abstract}

\section{Introduction}
Text retrieval is a fundamental task in information retrieval and plays a vital role in various language systems, including search ranking~\cite{nogueira2019passage} and open-domain question answering~\cite{chen2017reading}.  In recent years, the dual-encoder approach~\cite{lee2019latent,karpukhin2020dense}, which encodes queries/passages into vectors and matches them via the dot-product operation, has been the de-facto implementation. However,   this approach is limited by the embedding space bottleneck~\cite{lee2022contextualized} and missing fine-grained interaction~\cite{wang2022neural}.

\begin{figure}[t!]
\centering
  \includegraphics[width=1.0\linewidth]{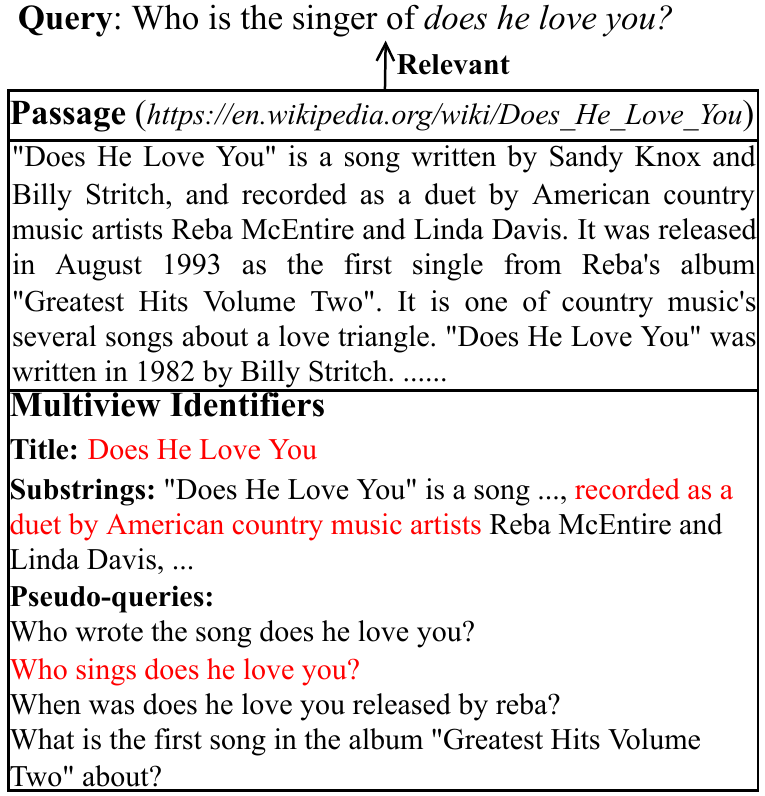}
  \vspace{-1.5em}
  \caption{An example of multiview identifiers for a passage. Corresponding to the query ``Who is the singer of does he love you?'', the semantic-related identifiers are highlighted in red.}
  \vspace{-1.5em}
  \label{example}
\end{figure}
An emerging alternative to the dual-encoder approach is generative retrieval~\cite{de2020autoregressive, tay2022transformer,bevilacqua2022autoregressive}. Generative retrieval utilizes autoregressive language models to generate identifier strings of passages, such as titles of Wikipedia pages, as an intermediate target for retrieval. The predicted identifiers are then mapped as ranked passages  in a one-to-one correspondence. Employing identifiers, rather than generating passages directly, could reduce useless information in a passage and makes it easier for the model to memorize and learn. At a cost, the identifier must be distinctive enough to represent a passage. Therefore, high-quality identifiers have been the secret to effective generative retrieval.

Previous studies have explored several types of identifiers, such as titles of documents~\cite{de2020autoregressive}, numeric IDs~\cite{tay2022transformer}, and distinctive substrings~\cite{bevilacqua2022autoregressive}. However, these identifiers are still limited: numeric IDs require extra memory steps and are ineffective in the large-scale corpus, while titles and substrings are only pieces of passages and thus lack contextualized information. More importantly, a passage should answer potential queries from different views, but one type of identifier only represents a passage from one perspective.

In this work, we argue that generative retrieval could be improved in the following ways:

(1) Synthetic identifiers. To address the limitations of titles and substrings in providing contextual information, we propose to create synthetic identifiers that are generated based on a passage's content.  In practice, we find the pseudo-queries, that are generated upon multiple segments of a passage, could serve as effective synthetic identifiers.  For example, as shown in Figure~\ref{example}, the pseudo-query "What is the first song in the album Greatest Hits Volume Two about?" spans multiple sentences in the passage. Once a query could be rephrased into a potentially-asked pseudo-query, the target passage could be effectively retrieved.

(2) Multiview identifiers.  We believe that a single type of identifier is not sufficient to effectively represent a passage. Using multiple types of identifiers, such as titles, substrings, and synthetic identifiers, can provide complementary information from different views. (i) One type of identifier, like the title, may be unavailable in some scenarios. In this case, synthetic identifiers could alternatively work.  (ii) Different views of identifiers are better suited for different types of queries. Titles could respond to general queries, while substrings are more effective for detailed ones. And the synthetic identifiers could cover some complex and difficult queries that require multiple segments. (iii) For one specific query, passages could be scored and ranked holistically from different views.

\begin{figure*}[t]
\centering
  \includegraphics[width=1.0\linewidth]{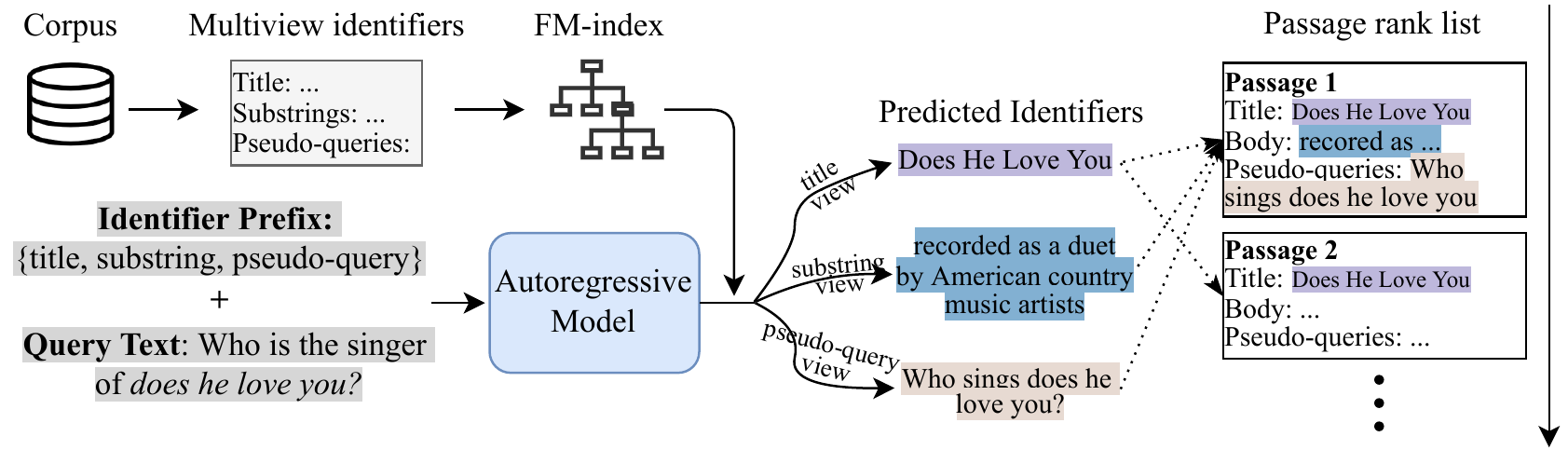}
  \vspace{-2em}
  \caption{Illustration of our proposed MINDER method. MINDER adopts multiview identifiers: the title, substrings, and pseudo-queries. For a query   with different identifier prefixes, MINDER generates corresponding identifiers in different views. Passages are ranked holistically according to the coverage with these generated identifiers. }
    \vspace{-1em}
  \label{method}
\end{figure*}

 Based on the above insights, we propose the {M}ultiview {I}dentifiers e{N}hance{D} g{E}nerative {R}etrieval approach, MINDER, as illustrated in Figure~\ref{method}. To represent a passage, we assign three views of identifiers: the title, substring, and synthetic identifiers (pseudo-queries). MINDER takes a query text and an identifier prefix indicating the type of identifier to be generated as input, and produces the corresponding identifier text as output.  Passages are ranked based on their coverage with the predicted three views of identifiers. We evaluate MINDER on three public datasets, and the experimental results show MINDER achieves the best performance among the current generative retrieval methods. 

The key contributions are summarized:
\vspace{-0.5em}
\begin{itemize}
\itemsep-0em 
  \item	We are the first to propose synthetic identifiers (generated based on the passage's content) to integrate contextualized information. In practice, we find pseudo-queries could serve as effective synthetic identifiers.  
  \item This is the first work that considers multiple views of identifiers simultaneously. Passages could be ranked holistically from different perspectives.
  \item Our approach achieves state-of-the-art performance in generative retrieval on three widely-used datasets.
\end{itemize}

\section{Related Work}
\subsection{Generative Retrieval}
Recently, we have witnessed an explosive development in autoregressive language models, such as the GPT-3/3.5 series~\cite{NEURIPS2020_1457c0d6,ouyang2022training}. This motivates the generative approach to retrieve passages. In some retrieval scenarios, like entity retrieval and sentence retrieval, the entire items could be regarded as identifiers. \citet{de2020autoregressive} proposed GENRE (Generative ENtity REtrieval), which retrieves an entity by generating the entity text itself. GENRE also could be applied in page-level retrieval, where each document contains a unique title as the identifier. \citet{lee2022generative} introduced generative retrieval to the multi-hop setting, and the retrieved items are short sentences. In 2022, \citet{tay2022transformer} proposed the DSI (Differentiable Search Index) method, which takes numeric IDs as identifiers for documents. \citet{wang2022neural} later improved the DSI by generating more queries as extra training data. However, the numeric Ids-based methods usually were evaluated on the small NQ320K datasets, partially because they suffer from the large scaling problem. \citet{bevilacqua2022autoregressive} proposed SEAL, which takes substrings as identifiers. The retrieval process is effectively completed upon the FM-Index structure. In this work, we mainly improve the SEAL method via synthetic identifiers and multiview identifiers. This is the first work that takes pseudo-queries as identifiers and considers multiple kinds of identifiers.
\subsection{Query Generation in Text Retrieval}
Query generation is originally introduced
to the IR community to improve the traditional term-based methods. \citet{nogueira2019doc2query} showed that appending the T5-generated
queries to the document before building the inverted index can bring substantial improvements
over BM25. More recently, \citet{mallia2021learning} used generated queries as term expansion to learn better sparse representations for documents.
In the context of dense retrieval, the generated pseudo-queries were used as extra data to improve the training process of dense retrieval. For example, \citet{ma2020zero} aimed to generate synthetic queries on the target domain for model training. \citet{dai2022promptagator} achieved excellent performance in few-shot retrieval with prompt enhanced query generation. In generative retrieval, \citet{wang2022neural} also explored the use of pseudo-queries as extra data to train DSI. In this paper, we are the first to use pseudo-queries as one view of identifiers for generative retrieval.
\subsection{Dense Retrieval}
In recent years, text retrieval has witnessed a paradigm shift from traditional BM25-based inverted index retrieval to neural dense retrieval~\cite{lee2019latent,karpukhin2020dense,li2022dynamic}. Dense retrieval is further developed via hard negative sample mining~\cite{xiong2020approximate,qu2021rocketqa} and  better pre-training design~\cite{chang2019pre,wang2022simlm}, and has achieved excellent performance. ~\citet{zhang2022multi} argued that a single vector representation of a document is
hard to match with multi-view queries and proposed the multi-view document representation vectors. This is similar to our work, but we focus on using multiview identifiers to improve generative retrieval.

Compared to dense retrieval that relies on the dual-encoder architecture, generative retrieval is promising to overcome the missing fine-grained interaction problem via the encoder-decoder paradigm. However, as a recently proposed technique route, generative retrieval still lags behind the state-of-the-art dense retrieval method and leaves much scope to investigate.

\section{Method}
Given a query text $q$, the retrieval system is required to retrieve a list of passages $p_1, p_2, \dots, p_n$, from a corpus $\mathcal{C}$. Both queries and passages are a sequence of text tokens. Besides, there are $k$
relevant query-passage pairs $\{q_i, p_i\}^k$ for training, where $p_i \in \mathcal{C}$.
\subsection{Multiview Identifiers}
For all passages in the corpus $\mathcal{C}$, we assign them multiview identifiers, including the titles, substrings, and pseudo-queries. These different types of identifiers could represent a passage from different perspectives.

\textbf{Title}. A title is usually a very short string that indicates the subject of a passage. Titles have been verified as effective identifiers in page-level retrieval. We denote a title as $t$ for a passage $p$ and select it as one view of identifiers in our work. 

\textbf{Substrings}. For a query, some substrings in the relevant passage are also semantically related. For example, for the query ``Who is the singer of does he love you?'' in Figure~\ref{example}, the substring ``recorded as a duet by'' is corresponding to the ``Who is the singer of'' in the query. For implementation, we directly store the whole content of the passage, denoted as $\mathcal{S}$, and sample substrings from $\mathcal{S}$ for model training.

\textbf{Pseudo-queries}. In this work, we generate pseudo-queries for a passage as synthetic identifiers to augment the title and substrings. Since pseudo-queries are generated based on the content of the passages, these synthetic identifiers could integrate multiple segments and contextualized information. For example, as shown in Figure~\ref{example}, the pseudo-query "What is the first song in the album
\textit{Greatest Hits Volume Two} about?" covers multiple sentences in the passage.

We first use the labeled query-passage pairs $\{q_i, p_i\}^k$ to train a query generation model $\textbf{QG}$. And then we generate a set of queries with top-k sampling strategy to encourage the query generation diversity. For each passage $p$ in corpus $\mathcal{C}$, we generate pseudo-queries $\mathcal{Q}$ as follows,
\begin{equation}  \label{eqn1}
   \begin{aligned}
   \mathcal{Q} = \textbf{QG}(p).
   \end{aligned}
 \end{equation} 

As such, for each passage in $\mathcal{C}$, we have obtained three views of identifiers $\{t, \mathcal{S}, \mathcal{Q} \}$. These identifiers could well represent a passage's content from different views.
\subsection{Model Training}
We train an autoregressive language model (denoted as $\textbf{AM}$) like BART~\cite{lewis2020bart} or T5~\cite{raffel2020exploring} to generate corresponding identifiers using the standard sequence-to-sequence loss. The input is the query text along with an identifier prefix, and the target is the corresponding identifier of the relevant passage, formulated as:
\begin{equation}  \label{eqn2}
   \begin{aligned}
   identifier = \textbf{AM}(prefix; q).
   \end{aligned}
 \end{equation}
The $prefix$ text is ``title'', ``substring'', and ``pseudo-query'', for the three different views, respectively. For the title view, the target text is the title $t$ of the relevant passage. For the substring view, we randomly select a substring $s$ from $\mathcal{S}$ as the target text. And to guarantee the semantic relevance between the input and the target, we only keep those substrings with a high character overlap with the query. As for the query view, we randomly select a pseudo-query $pq$ from $\mathcal{Q}$ as the target. Since both the user query $q$ and the pseudo-query $pq$ are conditioned on the same passage, they are usually about the same subject and even are different forms of the same question.
The three different training samples are randomly shuffled to train the autoregressive model.
\subsection{Model Inference}
In this section, we detail how to retrieve passages using the trained autoregressive model, $\textbf{AM}$.

\textbf{FM-index}. MINDER requires a data structure that can support generating valid identifiers. Following the work~\cite{bevilacqua2022autoregressive}, we use the FM-index~\cite{892127} to store all types of identifiers. For easy understanding, FM-index could be regarded as a special prefix tree that supports search from any position. Specifically, we flatten multiview identifiers into a sequence of tokens with special split tokens. For example, the identifiers of the passage in Figure~\ref{example} are flattened into ``<TS> Does He Love You <TE> Does He Love You is a song  written by Sandy Knox and Billy Stritch, and recorded as  ..., <QS> Who wrote the song does he love you? <QE> <QS> Who sings does he love you? ...'', where ``<TS>, <TE>, <QS>, <QE>'' are special tokens indicating the start and end of different types of identifiers. Given a start token or a string, FM-index could provide the list of possible token successors in $O(Vlog(V))$, where $V$ is the vocabulary size. Therefore, we could force the $\textbf{AM}$ model to generate valid identifiers.

\textbf{Constrained generation}. Upon the FM-index, MINDER could generate valid identifiers via constrained generation. For the title view, we input the prefix text ``title'' and query text into the $\textbf{AM}$ model, and force it to generate from the token ``<TS>''.  As such, MINDER could generate a set of valid titles via beam search, denoted as $\mathcal{T}_g$. For the substring view, the $\textbf{AM}$ model receives the prefix ``substring'' and query as input, and generates substrings $\mathcal{S}_g$ via constrained beam search. Similarly, the $\textbf{AM}$ model could generate valid pseudo-queries $\mathcal{Q}_g$ with the start token ``<QS>'' and end token ``<QE>''. We also save the language model scores for each generated text and utilize them in the following passage ranking stage. Notably, the language model score for a string is influenced by its length, which makes long strings, like pseudo-queries, have lower scores. Therefore, we add a biased score for the pseudo-query view to offset the influence.

\textbf{Passage ranking}. Previous generative retrieval methods~\cite{tay2022transformer, de2020autoregressive} could rank items directly using the constrained beam search, since their identifiers could map to passages one-to-one. Differently, MINDER considers multiview identifiers to rank passages comprehensively. To address this issue, we propose a novel scoring
formulation that aggregates the contributions of multiview identifiers. Each passage's score is holistically computed according to its coverage with the predicted identifiers, $\mathcal{T}_g$, $\mathcal{S}_g$, and $\mathcal{Q}_g$.

We follow the work~\cite{bevilacqua2022autoregressive} to rank passages with the generated identifiers. For a passage $p$, we select a subset $\mathcal{I}_p$ from the predicted identifiers. One identifier $i_p \in  \{\mathcal{T}_g$, $\mathcal{S}_g$, and $\mathcal{Q}_g\}$ is selected if $i_p$ occurs at least once in the identifiers of passage $p$. To avoid repeated scoring of substrings, we only consider once for substrings that overlapped with others. Finally, the rank score of the passage $p$ corresponding to the query $q$ is formulated as the sum of the scores of its covered identifiers,
\begin{equation}  \label{eqn3}
   \begin{aligned}
   s(q, p) = \sum_{i_p \in \mathcal{I}_p} s_{i_p},
   \end{aligned}
 \end{equation}
where $s_{i_p}$ is the language model score of the identifier $i_p$.

According to the rank score $s(q, p)$, we could obtain a rank list of passages from the corpus $\mathcal{C}$. In practice, we could use the FM-index to conveniently find those passages that contain at least one predicted identifier rather than score all of the passages in the corpus.

\begin{table*}[t]
\renewcommand\arraystretch{1}
  \centering
    \scalebox{1.0}{
    \begin{tabular}{cccccccc}
    \toprule
    \multicolumn{1}{c}{\multirow{2}*{Methods}}
    &\multicolumn{3}{c}{\makecell[c]{Natural Questions}}&&\multicolumn{3}{c}{\makecell[c]{TriviaQA}}\\\cline{2-4}\cline{6-8}
         &@5&@20&@100&&@5&@20&@100\cr
    \toprule
    BM25&43.6&62.9&78.1&&67.7&77.3&83.9\cr
    DPR\cite{karpukhin2020dense}&\textbf{68.3}&\textbf{80.1}&86.1&&\underline{72.7}&\underline{80.2}&\underline{84.8}\cr 
    GAR\cite{mao-etal-2021-generation}&59.3&73.9&85.0&&\textbf{73.1}&\textbf{80.4}&\textbf{85.7}\cr \toprule
    DSI-BART\cite{tay2022transformer}&28.3&47.3&65.5&&-&-&-\cr 
    SEAL-LM\cite{bevilacqua2022autoregressive}&40.5&60.2&73.1&&39.6&57.5&80.1\cr 
    SEAL-LM+FM\cite{bevilacqua2022autoregressive}&43.9&65.8&81.1&&38.4&56.6&80.1\cr 
    SEAL\cite{bevilacqua2022autoregressive}&61.3&76.2&\underline{86.3}&&66.8&77.6&84.6\cr 
    MINDER&$\underline{65.8}^{\dagger}$&$\underline{78.3}^{\dagger}$&$\textbf{86.7}^{\dagger}$&&$68.4^{\dagger}$&$78.1^{\dagger}$&$\underline{84.8}^{\dagger}$\cr \toprule
    \end{tabular}} 
    \vspace{-0.5em}
    \caption{ Retrieval performance on NQ and TriviaQA. We use hits@5, @20, and @100, to evaluate the retrieval performance. Inapplicable results are marked by “-”. The best results in each group are marked in Bold, while the second-best ones are underlined. \textbf{$\dagger$ denotes the best result in generative retrieval}.}  \label{tab:Retrieval performance}
    \vspace{-1em}
\end{table*}

\section{Experiments}
\subsection{Datasets}
We conducted experiments on widely-used NQ~\cite{kwiatkowski2019natural} and TriviaQA~\cite{joshi2017triviaqa} datasets with the DPR~\cite{karpukhin2020dense} setting. NQ and TriviaQA are open-domain QA datasets, where the queries are natural language questions and the passages are from Wikipedia. Each page in Wikipedia is chunked into several passages with no more than 100 words. Therefore, several passages may share the same Wikipedia title. Besides, we also evaluated generative retrieval methods on the MSMARCO dataset~\cite{nguyen2016ms}. MSMARCO is sourced from the Web search scenario, where queries are web search queries and passages are from Web pages. 

\subsection{Baselines}
We compared MINDER with the generative retrieval methods, DSI~\cite{tay2022transformer} and SEAL~\cite{bevilacqua2022autoregressive}. GENRE~\cite{de2020autoregressive} was excluded because it relies on unique titles of documents and thus cannot perform passage-level retrieval. Besides, we also included the term-based method, BM25, DPR~\cite{karpukhin2020dense},  and GAR~\cite{mao-etal-2021-generation} for comparison. Most of the results of baselines are from their paper, and the rest are reproduced by using publicly released code.

\subsection{Implementation Details}
For a fair comparison with previous work~\cite{bevilacqua2022autoregressive}, we utilized the BART-large as the backbone. 
We finetuned the model using training samples, title, substrings, and pseudo-queries, with the portion of 3:10:5. Inspired by SEAL that exposes the model to more possible pieces of evidence, we also add some “unsupervised” examples to the training set. In each of these examples, the model takes as input a random pseudo-query and generates the corresponding passage's identifiers. We discuss its influence in Section 4.7. 
~\citeauthor{lewis2021paq} have generated pseudo-queries for half of the passages on Wikipedia. Therefore, we generate queries for another half of the passages on Wikipedia. And for the MSMARCO corpus, we take the pseudo-queries from the work~\cite{nogueira2019doc2query}. 

We trained MINDER with the fairseq\footnote{\url{https://github.com/facebookresearch/fairseq}.} framework. We adopted the Adam optimizer with a learning rate of 3e-5, warming up for 500 updates, and training for 800k total updates. Detailed training hyperparameters are illustrated in Appendix~\ref{sec:Training_Hyperparameters} for better reproduction. The experiments are conducted on 8$\times$32GB NVIDIA V100 GPUs.

\subsection{Retrieval Results on QA}
The retrieval performance on NQ and TriviaQA is summarized in Table~\ref{tab:Retrieval performance}. By jointly analyzing the results, we gained the following findings.

(1) Among the generative retrieval methods, MINDER achieves the best performance. We found that SEAL which takes natural identifiers surpasses DSI based on numeric identifiers. This is because numeric identifiers lack semantic information and DSI requires the model to memorize the mapping from passages to their numeric IDs. As such, it becomes more challenging for DSI on the NQ and TriviaQA datasets with more than 20 million passages. Despite the superiority of SEAL, MINDER still outperforms it. Specifically, the improvements in terms of hits@5 are 4.5\% and 1.6\% on NQ and TriviaQA, respectively. This verifies the effectiveness of our proposed multiview identifiers, which could rank passages from different perspectives.

(2) On NQ, MINDER achieves the best performance in terms of hits@100 and the second-best results in terms of hits@5, 20. However, generative retrieval methods, including MINDER, perform worse than dual-encoder approaches on TriviaQA. Generative retrieval methods rely on the identifiers to represent passages, and cannot ``see'' the content of the passage. Although the QG module in our work generates pseudo-queries based on a passage's content, the autoregressive language model AM still cannot directly ``see'' the original content of the passage. Besides, autoregressive generation has the error accumulation problem. These are the disadvantages of generative retrieval and why it may not perform as well as dense retrievers in some scenarios.

\begin{table}[t]
\renewcommand\arraystretch{1}
  \centering
    \scalebox{0.9}{
    \begin{tabular}{ccccc}
    \toprule
    \multicolumn{1}{c}{\multirow{2}*{Methods}}
    &\multicolumn{4}{c}{\makecell[c]{MSMARCO}}\\\cline{2-5}
         &R@5&R@20&R@100&M@10\cr
    \toprule
    BM25&28.6&47.5&66.2&18.4\cr
SEAL&19.8&35.3&57.2&12.7\cr
    MINDER&\textbf{29.5}&\textbf{53.5}&\textbf{78.7}&\textbf{18.6}\cr \toprule
    only pseudo-query&24.9&48.9&72.5&15.5\cr
    only substring&18.7&38.7&64.9&11.5\cr
    only title&9.8&19.3&30.1&5.5\cr \toprule
    \end{tabular}}  
    \vspace{-0.5em}
    \caption{Retrieval performance on the MSMARCO dataset. R and M denote Recall and MRR, respectively. SEAL and MINDER are trained only with labeled query-passage pairs.}  \label{tab:search dataset}
    \vspace{-1em}
\end{table}

\subsection{Retrieval Results on Web Search}
Previous generative retrieval works~\cite{tay2022transformer, bevilacqua2022autoregressive} only verified the effectiveness on open-domain QA datasets, like NQ320k and NQ, but did not evaluate under the Web search scenario. To deeply analyze generative retrieval, we conducted experiments on the MSMARCO dataset and reported the results in Table~\ref{tab:search dataset}. Notably, we tried to implement DSI on MSMARCO but achieved poor performance. This may be due to the large-scaling problem of DSI, which requires a huge amount of GPU resources to work on a large-scale corpus. 

By analyzing the results in Table~\ref{tab:search dataset}, we found: 1) Different from the results on the QA datasets, SEAL performs worse than BM25 under the Web search scenario. Queries in Web search may only contain several keywords, which makes it hard for SEAL to learn the semantic correlation between queries and the substrings of passages. 2) MINDER surpasses SEAL and achieves a bigger performance improvement compared with the results on the QA datasets. This benefits from the multiview identifiers, which improve MINDER's robustness  under various scenarios. 3) MINDER outperforms BM25, particularly in terms of Recall@100. MINDER could recall passages from three different views, and thus achieves a better performance in Recall@100 than Recall@5.

\begin{table}[t]
\renewcommand\arraystretch{1}
  \centering
    \scalebox{1.0}{
    \begin{tabular}{cccc}
    \toprule
    \multicolumn{1}{c}{\multirow{2}*{Methods}}
    &\multicolumn{3}{c}{\makecell[c]{Natural Questions}}\\\cline{2-4}
         &@5&@20&@100\cr
    \toprule
    only query&59.0&72.5&80.9\cr
    only substring&60.2&74.3&84.5\cr
    only title&60.4&74.9&84.1\cr\toprule
    w/o pseudo-query&63.4&77.2&86.1\cr
    w/o substring&63.1&77.0&85.0\cr
    w/o title&63.9&76.6&85.3\cr\toprule
    MINDER&65.8&78.3&86.7\cr \toprule
    \end{tabular}}  
    \vspace{-0.5em}
    \caption{ Ablation study on different views of identifiers. We use ``w/o query'', ``w/o substrings'', and ``w/o title'' to respectively denote new models without considering the query flow, substrings, and title as identifiers. We also evaluate MINDER with only one view of the identifier.} 
    \label{tab:ablation study}
\end{table}

\begin{figure}[t!]
\centering
  \includegraphics[width=0.75\linewidth]{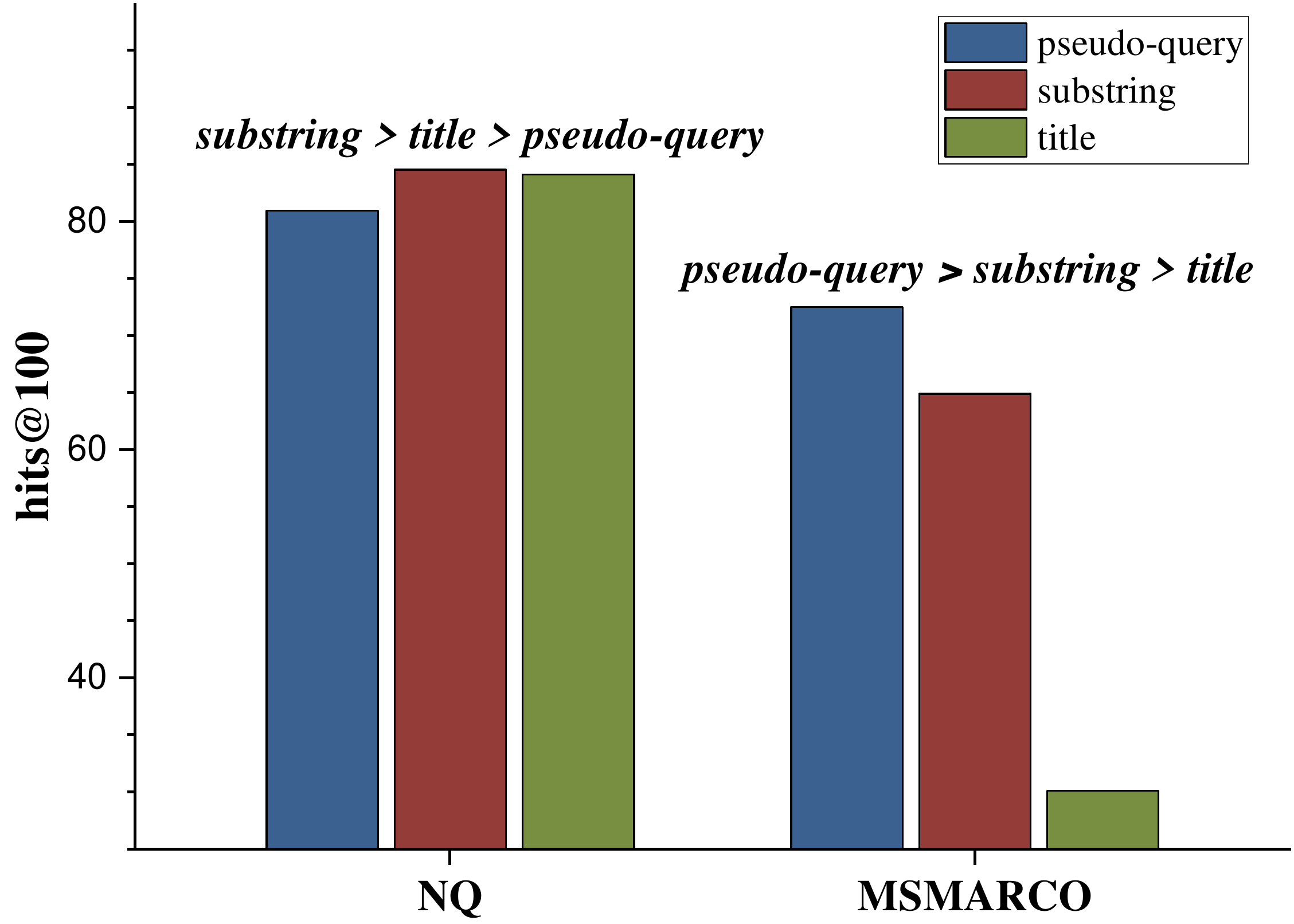}
  \vspace{-1em}
  \caption{Illustrating the roles of various identifier views in different search scenarios.}
  \label{view_comparison}
\end{figure}

\subsection{Ablation Study}
MINDER considers multiple types of identifiers: titles, substrings, and pseudo-queries. 1) Do the three views of identifiers all contribute to MINDER? 2) how much help does MINDER gain from the three different identifiers? 3) Is there any difference among different datasets? To answer these questions, we conducted experiments by eliminating one type of identifier each time. The results are illustrated in Table~\ref{tab:search dataset} and 
 Table~\ref{tab:ablation study}. To better demonstrate the functions of different views on different datasets, we kept only one view identifier and reported results in Figure~\ref{view_comparison}.

\begin{table}[t]
\renewcommand\arraystretch{1}
  \centering
    \scalebox{0.9}{
    \begin{tabular}{ccccc}
    \toprule
    \multicolumn{1}{c}{\multirow{2}*{Methods}}&\multicolumn{1}{c}{\multirow{2}*{\makecell[c]{Unsupervised \\data}}}
    &\multicolumn{3}{c}{\makecell[c]{Natural Questions}}\\\cline{3-5}
         &&@5&@20&@100\cr
    \toprule
        SEAL&\XSolidBrush&58.9&74.8&85.4\cr
    SEAL&span as queries&61.3&76.2&86.3\cr
    SEAL&pseudo-queries&61.2&76.8&85.7\cr \toprule
    MINDER&\XSolidBrush&64.6&76.8&86.4\cr
    MINDER&span as queries&65.9&78.3&86.7\cr
    MINDER&pseudo-queries&65.8&78.3&86.7\cr \toprule
    \end{tabular}} 
    \vspace{-0.5em}
    \caption{ Retrieval performance with different unsupervised data. ``span as queries'' and ``pseudo-queries'' means taking a span from the passage or a pseudo-query as the input, respectively.  }  \label{tab:unsupervised data}
\end{table}
\begin{table}[t!]
\renewcommand\arraystretch{1}
  \centering
    \scalebox{1.0}{
    \begin{tabular}{cccc}
    \toprule
    \multicolumn{1}{c}{\multirow{2}*{Methods}}
    &\multicolumn{3}{c}{\makecell[c]{Natural Questions}}\\\cline{2-4}
         &@5&@20&@100\cr
    \toprule
    MINDER+ID view&64.6&77.1&86.1\cr
    MINDER&64.6&76.8&86.4\cr \toprule
    \end{tabular}}  
     \vspace{-0.5em}
    \caption{ Evaluation of numeric identifiers as one view identifier in MINDER. Both two variants are trained only with labeled query-passage pairs.}  \label{tab:numeric identifiers}
    \vspace{-1em}
\end{table}
\begin{table}[t]
\renewcommand\arraystretch{1}
  \centering
    \scalebox{1.0}{
    \begin{tabular}{ccccc}
    \toprule
    &\multicolumn{1}{c}{\multirow{1}*{BS}}
         &@5&@20&@100\cr
    \toprule
    \multirow{4}*{TriviaQA}&5&66.9&77.1&83.8\cr
    &10&67.8&77.9&84.6\cr
    &15&68.4&78.1&84.8\cr
    &20&68.4&78.4&84.8\cr\toprule
    \multirow{4}*{\makecell[c]{MS\\MARCO}}&5&29.4&52.9&78.4\cr
    &10&29.4&53.9&79.3\cr
    &15&29.1&53.7&79.6\cr
    &20&27.8&52.8&79.8\cr\toprule
    \end{tabular}} 
    \vspace{-0.5em}
    \caption{ Retrieval performance of MINDER with beam size values in \{5, 10, 15, 20\}.}  \label{tab:beam size}
    \vspace{-1em}
\end{table}

From the results, we gained the following insights. (1) No matter which view of identifiers is removed from MINDER, the performance significantly declines. In terms of hits@5, the decline is 2.4\%, 2.7\%, and 1.9\%, while eliminating the pseudo-query view, substring view, and title view, respectively.  This clearly reveals that all three views of identifiers contribute to the system's performance, and verifies the necessity to adopt multiview identifiers simultaneously. (2) Besides, comparing the three types of identifiers, we found that eliminating the substring view degrades the most on NQ. This may be due to the fact that the substrings could cover the most content of a passage. Although the ``only title'' and ``only pseudo-query'' variants perform worse than the substring view, they could complement each other and significantly improve the overall performance. 
3) Comparing the results on NQ and MSMARCO, we found different views played different roles in different search scenarios. As illustrated in Figure~\ref{view_comparison}, the substring view is vital on NQ while the pseudo-view contributes the most on MSMARCO. This is determined by the different natures between the QA and Web search scenarios. And it verifies the necessity to adopt multiview identifiers again.

\begin{figure*}[t]
  \centering
  \subfigure{
  \includegraphics[width=1.0\linewidth]{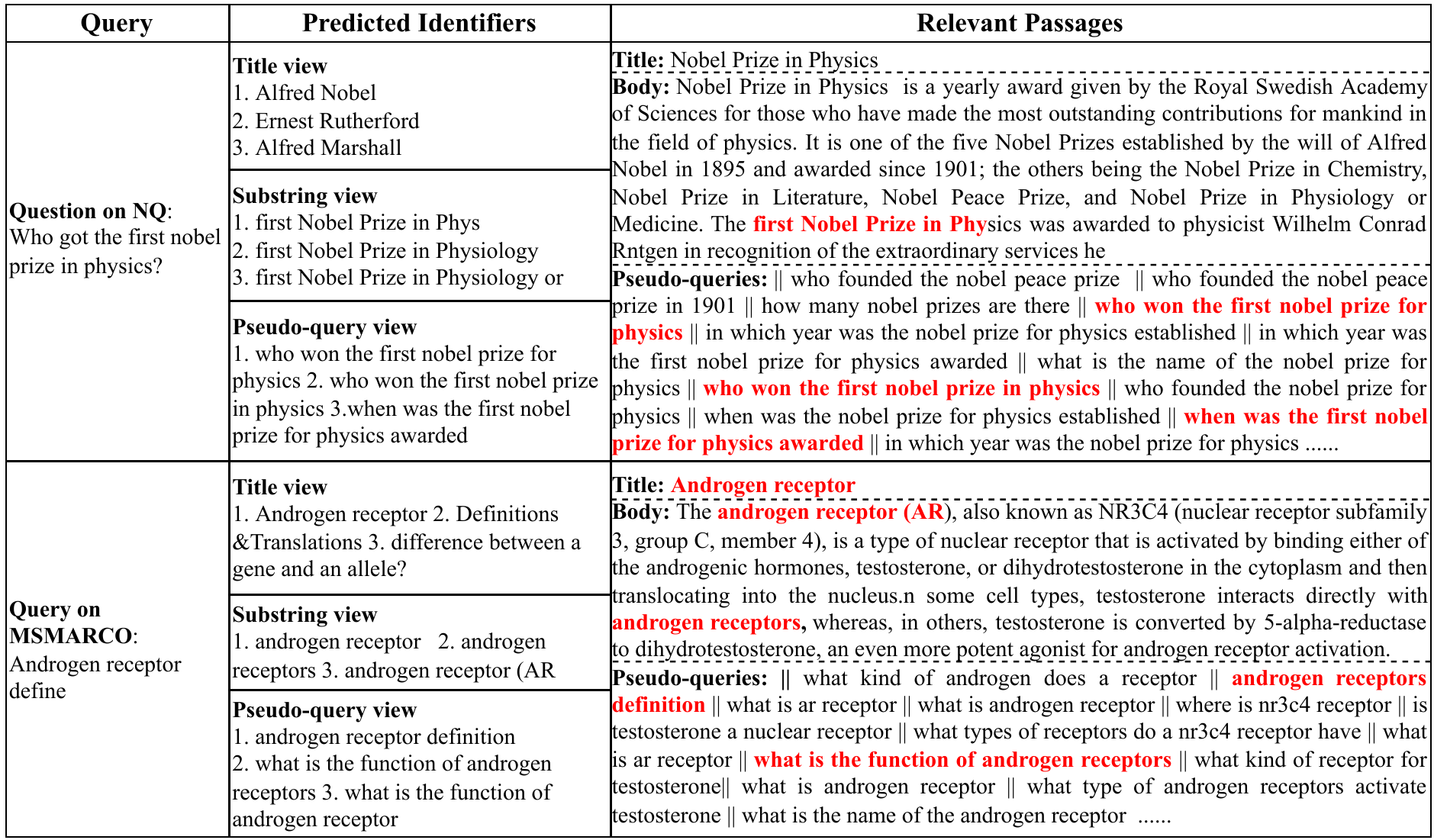}
   }
   \vspace{-1.5em}
  \caption{Case study. Two cases from NQ and MSMARCO. For the predicted identifiers from MINDER, we show three top-scored predictions for the title view, body view, and pseudo-query view, respectively. The predicted identifiers that occur in relevant passages are colored in red.}\label{case_study}
   \vspace{-1em}
\end{figure*}

\subsection{In-depth Analysis}
\textbf{Unsupervised Data}. Besides the labeled query-passage pairs, we also trained MINDER using pseudo-queries. SEAL conducted unsupervised data by randomly selecting a span from a passage as the input. (1) Are the unsupervised data useful for the training? (2) Which kinds of unsupervised data contribute most? We conducted experiments by using different kinds of unsupervised data, and the results are illustrated in Table~\ref{tab:unsupervised data}. We found that both kinds of unsupervised data improve upon purely supervised training. Specifically, the performance gets improved by 2.3 and 1.2 points in terms of hits@5 for SEAL and MINDER respectively. There is no significant gap between the two kinds of unsupervised data. We think the unsupervised training mainly exposes passages to the model, and both two ways could meet this goal.

\textbf{Numeric Identifiers}. MINDER adopts multiview identifiers, including titles, substrings, and pseudo-queries, which are all semantic text. We excluded  numeric identifiers in MINDER, because IDs are numbers and lack semantic information. As such, numeric identifiers require extra steps to memorize the mapping from passages to IDs. For exploration, we also added the ID view in MINDER and reported the results in Table~\ref{tab:numeric identifiers}. It is observed that there is no big difference in performance after including numeric identifiers. On the one hand, numeric identifiers are weak at large-scale corpus. Therefore, the ID view cannot contribute to MINDER on the NQ dataset. On the other hand, numeric identifiers fail to provide extra information to complement the three views identifiers in MINDER.

\textbf{Beam Size}. MINDER relies on beam search to predict a set of identifiers, and then these predicted identifiers are mapped as ranked passages. To evaluate the influence of beam size, we conducted experiments and reported results in Table~\ref{tab:beam size}. The results suggest that a bigger beam size, like 15 or 20, could achieve a better performance in terms of hits@100 on both two datasets. As for the top-ranked evaluation, TriviaQA prefers a bigger beam size, but MSMARCO requires a smaller one. One possible reason is that there are too many similar passages on MSMARCO and a bigger beam size introduces more noise.

\textbf{Inference speed}. On our equipment, MINDER takes about
135 minutes to complete the inference process on the NQ test set, while SEAL takes about 115 minutes. Both of them apply the same beam size of 15. MINDER requires
1.2 times more inference time than SEAL on our equipment, due to the increased identifier views. 

\subsection{Case Study}
To qualitatively illustrate why MINDER works, we analyzed the prediction results on NQ and MSMARCO in Figure~\ref{case_study}. (1) It is observed that pseudo-queries are sufficient and could cover almost potential queries. In the first example, given the question ``Who got the first nobel prize in physics?'', MINDER generates either the same meaning question ``who won the first nobel prize for
physics'' or another question about the same subject ``when was the first novel prize for physics award''. These predicted queries accurately locate the relevant passage. (2) As for the substring view, MINDER tends to generate almost the same ones. These substrings are not much distinctive and could be found in several passages of the corpus. This may be the reason why the substring view cannot work well on MSMARCO. 
\section{Conclusion and Future Work}
In this work, we present MINDER, a novel retrieval system that combines an autoregressive language
model with multiview identifiers. We find pseudo-queries are admirable identifiers that could work on different search scenarios. More importantly, MINDER simultaneously utilizes multiple types of identifiers, including titles, substrings, and pseudo-queries. These different views of identifiers could complement each other, which makes MINDER effective and robust in different search scenarios. The experiments on three widely-used datasets illustrate MINDER achieves the best performance in generative retrieval.

In the future, we aim to improve MINDER from the following aspects.MINDER adopts a heuristic function to aggregate predicted identifiers and rank passages. The heuristic rank function relies on manual hyper-parameters to balance different views of identifiers, which may not be suitable for all samples. As such, we are motivated to integrate the rank process into an auto-learned neural network. Besides, we plan to apply MINDER on more search domains, like the few-shot retrieval setting.
\section*{Acknowledgments}
The work described in this paper was supported by Research Grants Council of Hong Kong(PolyU/5210919, PolyU/15207821, and PolyU/15207122), National Natural Science Foundation of China (62076212) and PolyU internal grants (ZVQ0).

\section*{Limitations}
MINDER achieves the best performance among the current generative retrieval methods, but it is still not as good as the well-designed dual-encoder approaches and lags behind the current state-of-the-art on leaderboards. The reason for this is that the model's autoregressive generation way (generating from left to right) prevents it from "seeing" the entire content of a passage. Generative retrieval methods have advantages over dual-encoder approaches but also leave many research problems to be investigated. Another limitation of MINDER is the memory consumption of identifiers. Since MINDER considers multiview identifiers, it also consumes more memory to store these identifiers. Fortunately, we use the FM-index structure to process the identifiers, and the space requirements are linear in the size of the identifiers.
\section*{Ethics Statement}
The datasets used in our experiment are publicly released and labeled through interaction with humans in English. In this process, user privacy is protected, and no personal information is contained in the dataset. The scientific artifacts that we used are available for research with permissive licenses. And the use of these artifacts in this paper is consistent with their intended use. Therefore, we believe that our research work meets the ethics of ACL. 
\bibliography{anthology,custom}
\bibliographystyle{acl_natbib}

\appendix
\section{Training Hyperparameters}
\label{sec:Training_Hyperparameters}
\begin{table}[ht]
\centering
 \scalebox{0.85}{
\begin{tabular}{cc}
\toprule
{Name}&{Value}\\
\toprule
{arch}&bart\_large\\\cline{1-2}
{task}&translation\\\cline{1-2}
{criterion}&label\_smoothed\_cross\_entropy\\\cline{1-2}
{weight-decay}&0.01\\\cline{1-2}
{optimizer}&adam\\\cline{1-2}
{lr-scheduler}&polynomial\_decay\\\cline{1-2}
{lr}&3e-05 \\\cline{1-2}
{total-num-update}&$800000$ \\\cline{1-2}
{patience}&5 \\
\bottomrule
\end{tabular}}
\vspace{-0.5em}
\caption{Hyperparameters to train MINDER using the fairseq.}
\label{tab:Training_Hyperparameters}
\end{table}

For better reproduction, we detail the training hyperparameters in Table~\ref{tab:Training_Hyperparameters}. We train our model for serval runs with the fairseq, and the results of the different runs are reported in Table~\ref{tab:multi-runs}.

\begin{table}[h]
\renewcommand\arraystretch{1}
  \centering
    \scalebox{1.0}{
    \begin{tabular}{cccc}
    \toprule
    \multicolumn{1}{c}{\multirow{2}*{\# Run }}
    &\multicolumn{3}{c}{\makecell[c]{Natural Questions}}\\\cline{2-4}
         &@5&@20&@100\cr
    \toprule
    1&66.2&78.6&86.9\cr
    2&66.2&78.6&86.9\cr
    3&65.8&78.3&86.7\cr
    4&64.8&78.6&86.7\cr\toprule
    \end{tabular}}  
    \vspace{-0.5em}
    \caption{ Results of MINDER on NQ for different runs.} 
    \label{tab:multi-runs}
\end{table}

\end{document}